\documentclass[journal]{IEEEtran}
\usepackage{algorithm}
\usepackage{multirow}
\usepackage{array}
\usepackage{algpseudocode}
\usepackage{amsmath}
\usepackage{cite}
\usepackage{booktabs}
\usepackage{graphicx}
\usepackage{float}
\usepackage{threeparttable}
\usepackage{epstopdf}
\usepackage{xfrac}
\usepackage{makecell}
\usepackage{amssymb}
\usepackage{subfigure}
\usepackage{bm}
\usepackage{cases}
\usepackage{url}
\usepackage{hyperref}

\newcommand{\Rmnum}[1]{\expandafter\@slowromancap\romannumeral #1@}
\makeatother

\begin{document}

\title{Evolutionary Biparty Multiobjective UAV Path Planning: Problems and Empirical Comparisons}

\author{Kesheng~Chen, Wenjian~Luo,~\IEEEmembership{Senior Member,~IEEE,} Xin~Lin, Zhen~Song, Yatong~Chang
	\thanks{This study is supported by the Shenzhen Fundamental Research Program (Grant No. JCYJ20220818102414030), the Major Key Project of PCL (Grant No. PCL2022A03), the Shenzhen Science and Technology Program (Grant No. ZDSYS20210623091809029), and the Guangdong Provincial Key Laboratory of Novel Security Intelligence Technologies (Grant No. 2022B1212010005). \textit{(Corresponding author: Wenjian Luo.)}}
	\thanks{Kesheng~Chen, Wenjian~Luo, Zhen~Song, and Yatong~Chang are with Guangdong Provincial Key Laboratory of Novel Security Intelligence Technologies, School of Computer Science and Technology, Harbin Institute of Technology, Shenzhen 518055, China. Wenjian Luo is also with Peng Cheng Laboratory, Shenzhen 518055, China. Xin~Lin is with the School of Artificial Intelligence/School of Future Technology, Nanjing University of Information Science and Technology, Nanjing 210044, China. (E-mail: 22s151138@stu.hit.edu.cn, luowenjian@hit.edu.cn, xlin@nuist.edu.cn, 21s151097@stu.hit.edu.cn, 20s151150@stu.hit.edu.cn.)}}

\maketitle
\begin{abstract}
Unmanned aerial vehicles (UAVs) have been widely used in urban missions, and proper planning of UAV paths can improve mission efficiency while reducing the risk of potential third-party impact. Existing work has considered all efficiency and safety objectives for a single decision-maker (DM) and regarded this as a multiobjective optimization problem (MOP). However, there is usually not a single DM but two DMs, i.e., an efficiency DM and a safety DM, and the DMs are only concerned with their respective objectives. The final decision is made based on the solutions of both DMs. In this paper, for the first time, biparty multiobjective UAV path planning (BPMO-UAVPP) problems involving both efficiency and safety departments are modeled. The existing multiobjective immune algorithm with nondominated neighbor-based selection (NNIA), the hybrid evolutionary framework for the multiobjective immune algorithm (HEIA), and the adaptive immune-inspired multiobjective algorithm (AIMA) are modified for solving the BPMO-UAVPP problem, and then biparty multiobjective optimization algorithms, including the BPNNIA, BPHEIA, and BPAIMA, are proposed and comprehensively compared with traditional multiobjective evolutionary algorithms and typical multiparty multiobjective evolutionary algorithms (i.e., OptMPNDS and OptMPNDS2). The experimental results show that BPAIMA performs better than ordinary multiobjective evolutionary algorithms such as NSGA-II and multiparty multiobjective evolutionary algorithms such as OptMPNDS, OptMPNDS2, BPNNIA and BPHEIA.
\end{abstract}

\IEEEpeerreviewmaketitle

	\section{Introduction}
\label{sec:intro}
\IEEEPARstart{U}{AVs} have been widely used in urban environments for aerial photography \cite{kwong2020tree}, unmanned deliveries \cite{torija2021psychoacoustic}, urban traffic monitoring \cite{butilua2022urban}, infrastructure maintenance \cite{outay2020applications}, and many other operations. Studies show that the use of UAVs for daily operations in metropolitan areas will become mainstream in the future, and government departments are beginning to regulate UAV operations from the perspectives of public safety, property safety, and environmental factors \cite{narkus-kramerFutureDemandBenefits2017}.

To date, much work has adopted the formulation of multiobjective optimization problems (MOPs) to describe UAV path planning problems. For example, Peng\emph{ et al.} \cite{pengConstrainedMultiObjectiveOptimization2022,pengDecompositionbasedConstrainedMultiobjective2022} considered this a biobjective optimization problem that minimizes the flight path and the threat level and modeled the UAV path planning problem as a constrained multiobjective optimization problem, whose objectives are the route length and route altitude and whose constraints are terrain, turning angle, elevation, slope angle, and altitude. In \cite{ sunEvolutionaryAlgorithmConstraint2022,sunPathPlanningGEOUAV2016}, Sun\emph{ et al.} conducted experiments using a biobjective optimization model with a combination of imaging quality, path length and safety indicators. Zhang\emph{ et al.} \cite{zhangMultiobjectiveParticleSwarm2022} took efficiency and safety as one objective and weighted map constraints, terrain constraints and other factors as another objective. Besada-Portas\emph{ et al.} \cite{besada-portasEvolutionaryTrajectoryPlanner2010} summarized the objectives of UAV path planning efficiency in recent years, including the objectives of path length, path height, and path energy consumption.



In existing work on UAV path planning with multiple objectives, evolutionary algorithms (EAs) are a class of efficient techniques \cite{aitsaadiUAVPathPlanning2022a}. For example, Pang\emph{ et al.} \cite{pangUAVPathOptimization2022a} used a decomposition-based evolutionary algorithm with the constraint handling technique to solve multiobjective UAV path optimization problems. Sun\emph{ et al.} \cite{sunEvolutionaryAlgorithmConstraint2022} used a multiobjective evolutionary algorithm based on decomposition with the constraint relaxation technique. Zhang\emph{ et al.} \cite{zhangMultiobjectiveParticleSwarm2022} used multiobjective particle swarm optimization (MOPSO) combined with reinforcement learning, while Ghambari\emph{ et al.} \cite{ghambariEnhancedNSGAIIMultiobjective2020} used an improved nondominated sorting genetic algorithm II (NSGA-II). In addition, many works have used coevolutionary algorithms to study this problem. Zhu\emph{ et al.} \cite{zhu2022ucav} used a coevolutionary swarm intelligence algorithm to study UAV path planning. Xu\emph{ et al.} \cite{xu2023cooperative} considered coevolutionary search under the presence of multiple UAVs.

However, these works only consider UAV path planning (UAVPP) with a single DM model and use ordinal MOEAs to solve the corresponding model. Although modeling the efficiency objectives and safety objectives of the UAVPP problem as a MOP can enable a certain degree of balance between safety and efficiency, there are still some significant drawbacks. The UAVPP model with a single DM ignores the connection between optimization objectives and multiple DMs, which makes it difficult to obtain common Pareto optimal solutions from the perspective of different DMs. Thus, the solution set of this model is not guaranteed to be Pareto optimal from the perspectives of the efficiency DM and safety DM.


In this paper, we establish a biparty multiobjective UAV path planning (BP-UAVPP) model with the efficiency DM and safety DM to precisely describe practical requirements. The efficiency DM tends to minimize efficiency factors such as flight time and flight energy consumption, while the safety DM tends to minimize risk factors such as personal risk and noise pollution.

The UAV path planning problem with two decision-makers considered in this paper is a biparty multiobjective optimization problem (BPMOP). BPMOPs are MOPs with two decision-makers, and more generally, MOPs with multiple decision-makers are called multiparty multiobjective optimization problems 
(MPMOPs). 
The goal of MPMOPs is to search for optimal solutions as close to the Pareto front of each DM as possible. Multiobjective evolutionary algorithms have difficulty solving BPMOPs directly \cite{liuEvolutionaryApproachMultiparty2020,10.1007/978-3-030-78811-7_6,chang2022multiparty,song2022multiobjective}. To better illustrate this point, the following is a practical example of solving a BP-UAVPP.

\begin{figure}[h]
	\centering  
	\includegraphics[scale=0.7,trim=150 200 200 200,clip]{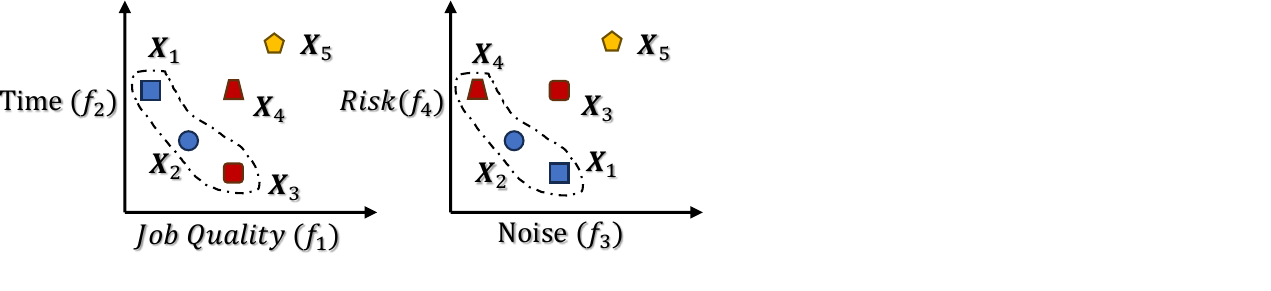}    
	\caption{Each solution has four objectives and solution sets from the perspectives of different DMs.}
	\label{fig8}
\end{figure}

Fig. \ref{fig8} shows a simplified example of Pareto optimal solutions to an ordinary multiobjective UAV path planning problem with efficiency-related and safety-related objectives. There are four objectives: the efficiency DM's objectives are job quality and time consumption, and the safety DM's objectives are fatality risk and noise pollution. In an ordinary MOP considering all objective functions together, solutions $X_1$-$X_4$ constitute a Pareto optimal set (PS). However, from the perspective of a single DM, they are not necessarily Pareto optimal. In Fig. \ref{fig8}, markers of the same shape indicate identical solutions. The red solutions $X_3, X_4$ are Pareto optimal solutions from the perspective of one DM but not Pareto optimal from the perspective of another DM. In contrast, the blue solutions are Pareto optimal from the perspectives of all DMs. The blue solutions $X_1, X_2$ are more valuable in practice than the red solutions $X_3, X_4$.

\begin{table}[H]
	\centering
	\caption{Differences in solutions under different algorithms}
	\begin{tabular}{cccc}
		\toprule[1.5pt]
		Solution &  $f_1-f_4$ & NSGA-II layers & OptMPNDS2 layers \cite{10.1007/978-3-030-78811-7_6} \\
		\cline{1-4}
		$X_1$ & (1,3,3,1) & 1 & 1 \\
		$X_2$ & (2,2,2,2) & 1 & 1 \\
		$X_3$ & (3,1,3,3) & 1 & 2 \\
		$X_4$ & (3,3,1,3) & 1 & 2 \\
		$X_5$ & (4,4,4,4) & 2 & 3 \\
		\bottomrule[1.5pt]
	\end{tabular}
	\label{SMALL CASE}    
\end{table}

For clarity, Table \ref{SMALL CASE} shows the detailed objective function values of the five solutions and the numbers of ranking layers under different algorithms. If we consider only the ordinary nondominated sorting in the MOP, the numbers of ranking layers of X1-X4 are all 1. The algorithm eventually retains X1-X4, and all these solutions are consistent in terms of convergence. However, in the biparty multiobjective optimization problem, compared with X3 and X4, X1 and X2 are closer to the frontier of both DMs. In fact, the rankings of $\{ x_1,x_2\}$ and $\{ x_3,x_4\}$ are different in OptMPNDS2 \cite{sheNewEvolutionaryApproach2021}.

Therefore, the ordinary multiobjective optimization algorithm is not able to distinguish this situation effectively, which ultimately means that the solution results do not satisfy the needs of BPMOPs. The optimal solution set with two or more decision-makers is a small part of the optimal solution set obtained when treating all objectives as a single DM. However, it is possible to obtain a larger solution set by solving the UAVPP with a single DM model and finally filtering the subset of solutions from the solution set that lies on the common frontier of the safety and efficiency decision sides as a solution. However, this approach fails to incorporate selection pressure into the algorithm's search process, making the algorithm inefficient in solving such problems.

In addition, there are currently only a few multiparty multiobjective evolutionary algorithms (MPMOEAs) based on the NSGA-II framework, such as OptMPNDS \cite{liuEvolutionaryApproachMultiparty2020} and OptMPNDS2 \cite{10.1007/978-3-030-78811-7_6}, to better solve and test the BP-UAVPP. This paper redesigns some operations of NNIA \cite{gongMultiobjectiveImmuneAlgorithm2008}, HEIA \cite{linHybridEvolutionaryImmune2015}, and AIMA \cite{linAdaptiveImmuneinspiredMultiobjective2018} based on the multiobjective immune algorithm framework to better solve BPMOPs or MPMOPs, and these methods are named BPNNIA, BPHEIA, and BPAIMA.

In summary, the contributions of this paper are summarized as follows.
\begin{enumerate}
	\item For the first time, a biparty multiobjective UAV path planning model in an urban operating environment is established. The model has two parts. The first part involves efficiency, and it minimizes objectives such as the UAV planning path length, height variation, fuel consumption and mission hover point distance to achieve high efficiency. The other involves safety, and it minimizes objectives such as fatality risk, noise pollution, and urban property safety to achieve high safety.
	\item The multiobjective
	immune algorithm with nondominated neighbor-based
	selection (NNIA) \cite{gongMultiobjectiveImmuneAlgorithm2008}, the hybrid evolutionary framework for multiobjective immune algorithms (HEIA) \cite{linHybridEvolutionaryImmune2015}, and the adaptive immune-inspired multiobjective algorithm (AIMA) \cite{linAdaptiveImmuneinspiredMultiobjective2018} are adapted to obtain BPNNIA, BPHEIA and BPAIMA to solve biparty multiobjective UAV path planning problems. These algorithms are compared with NSGA-II and typical MPMOEAs. The experimental results on the biparty UAV path scheduling problem show that these algorithms perform better than typical MOEAs and MPMOEAs.
\end{enumerate}

The remaining sections of this paper are organized as follows. Section \ref{sec:backgrounds} introduces the concepts of MOPs and MPMOPs. Section \ref{sec:mpmuav} models the multiobjective UAV path planning problem as a biparty multiobjective optimization problem. Section \ref{sec:design} describes the design of the experiment, including the compared algorithms and the metric. In Section \ref{sec:ex}, the experimental results and analysis are given. Finally, a brief conclusion is given in Section \ref{sec:con}.

\section{Background} 
\label{sec:backgrounds}
In this section, multiobjective optimization problems and multiparty multiobjective optimization problems are briefly reviewed.

\subsection{Multiobjective Optimization Problems} 
A multiobjective optimization problem (MOP) \cite{debFastElitistMultiobjective2002a} is considered a problem with multiple conflicting objectives. Without loss of generality, a minimized MOP can be defined as follows.
\begin{equation}
	\begin{aligned}
		& \min F(X)=(f_{1}(X),f_2(X),\dots,f_m(X)),\\
		&	S.t. \begin{cases}
			g_i(X) \leq 0, &  i = 1,\dots,n_g \\
			h_j(X) = 0   , &  j = n_g+1,\dots,n_g+n_h \\
			X \in \Omega& 
			\end{cases} \\
	\end{aligned}
\end{equation}
where $g_i(X)$ denotes the $i$th inequality constraint, $h_j(X)$ denotes the $j$th equality constraint on $X$, and $n_g$ and $n_h$ are the numbers of inequality constraints and equality constraints, respectively. $X=(x_1,x_2,\dots,x_d)$ is an $d$-dimensional decision variable. The parameter $m$ denotes the number of objectives, $f_i(X)$ is the $i$th objective function, and $\Omega$ denotes the decision space. 

For two decision vectors $X$ and $Y$, if none of the objectives of $X$ are greater than the objectives of $Y$ and there exists at least one objective of $X$ that is less than the corresponding objective of $Y$, it can be said that $X$ Pareto dominates $Y$, which is denoted as $X \prec Y $. Based on the definition of Pareto dominance, the solution $X^P$ is Pareto optimal when there are no other Pareto dominating solutions $X^P$. In addition, all Pareto optimal solutions compose the Pareto optimal set (PS), and the corresponding objective vectors are called the Pareto front (PF).

\subsection{Multiparty Multiobjective Optimization Problems}
The multiparty multiobjective problem (MPMOP) was recently proposed by Liu\emph{ et al.} \cite{liuEvolutionaryApproachMultiparty2020}; it has multiple DMs, each DM represents a department, and at least one DM is concerned with multiple objectives. The objective set of the $k$th DM is $M_k$, and the objective values of individual $x$ for the $k$th decision-maker can be denoted as $F_{k}(X) =\{f_{k,i}(X)|f_{k,i}(X) \in M_k\}$. An MPMOP that minimizes all objective functions can be described as follows.
\begin{equation}
	\begin{aligned}
		& \min E(X)=(F_{1}(X),F_2(X),\dots,F_K(X)),\\
		&	where \begin{cases}
			F_{1}(X)=\{f_{11}(X),f_{12}(X),\dots,f_{1m_{1}}(X)\}\\
			F_{2}(X)=\{f_{21}(X),f_{22}(X),\dots,f_{2m_{2}}(X)\}\\
			\ \ \ \ \vdots\\
			F_{K}(X)=\{f_{K1}(X),f_{K2}(X),\dots,f_{Km_{K}}(X)\}\\
		\end{cases} \\
	\end{aligned},	
\end{equation}
where $E$ is the set of objective vectors from the $K$ DMs in the MPMOP. $f_{k,i}$ is the $i-th$ objective of the $k-th$ DM. The parameter $m_k$ denotes the number of objectives considered by the $k$th decision-maker.

We use $X \prec_k Y$ to denote that $X$ Pareto dominates $Y$ in the objective space $F_{k}(X)$ of the $k$th decision-maker. The objective of solving an MPMOP is to search for the solutions that are as close to the Pareto frontier in the objective space of each DM as possible. In a scenario with multiple DMs, the demands of each DM are often different to some extent. If an MPMOP is simply considered a MOP for optimization, some Pareto optimal solutions may be dominated by other solutions from the perspective of a single DM (the situation mentioned in Fig. \ref{fig8} is a typical example). The difference between the PS of a MOP and the PS of an MPMOP means that different algorithms are required. Therefore, it is necessary to use a well-designed algorithm to solve MPMOPs.

\section{Biparty Multiobjective UAV Path Planning}
\label{sec:mpmuav}
For the first time, this paper establishes a biparty multiobjective UAV path planning (BP-UAVPP) model with an efficiency DM and safety DM to precisely describe practical requirements. The BP-UAVPP model is compared to the general single-decision-maker UAV path planning model, which allows the problem to better express the real-world UAV path planning problem 
that exists for large numbers of UAVs 
when the efficiency decision-maker and the regulator make decisions together. The efficiency department is concerned with efficiency-related objectives such as the path length, path height, flight energy consumption, and mission hover point distance, while the safety department is concerned with third-party risk-related objectives such as the fatality risk, the property risk, and noise pollution.

\subsection{Efficiency Decision-Maker}\label{eff}

For the efficiency DM, maximizing the efficiency of the UAV flight is his or her core demand. This objective is mainly reflected in the objectives of minimizing the length of the UAV trajectory, minimizing the flight altitude changes, minimizing the flight energy consumption, and minimizing the total sum of the mission hover point distances. For a given path, the $i$th discrete trajectory points can be expressed as $ \vec{p_i}=(x_i,y_i,z_i)$, and the $i$th track segment can be expressed as $g_i=\vec{p}_{i+1}-\vec{p}_{i}$.
\subsubsection{Path length \textup{\cite{pengDecompositionbasedConstrainedMultiobjective2022}}} 
In the process of a UAV mission, the planned path length of the UAV directly affects the time of UAV operation, and users who care about efficiency often want to choose the shortest path to reduce the operation time. In this paper, the path length of all UAV segments is selected as the objective function, which is shown as follows.
\begin{equation}
	f_{\textup{length}} = \sum_{i=0}^{n-1}||g_i|| .
\end{equation}
\subsubsection{Flight energy consumption \textup{\cite{ghambariEnhancedNSGAIIMultiobjective2020}}}
In the $i$th track segment, the flight energy consumption of a small-rotor UAV mainly depends on the flight state, hovering power consumption, and energy of overcoming gravity. The power consumption formula $fuel_i$ in the $i-th$ track segment is given as follows.
\begin{equation}
	fuel_i = W^{\frac{3}{2}}\sqrt{\frac{G^3}{2\rho_i \zeta n}}\frac{||g_i||}{V}+\Delta^{+}WG,
\end{equation}
where the parameter $W$ is the weight of the UAV and the battery, $\rho$ is the fluid density of air, $G$ is gravity, $\Delta^{+}$ is the altitude increase, $n$ is the number of rotors, $||g_i||$ is the track length, $V$ is the flight speed, and $\zeta $ is the area of the rotating blade disk. The atmospheric density as a function of altitude can be expressed as $\rho_i=\rho_0\exp(-(z_{i+1}+z_{i})/(2*10.7)),$
where $\rho_0$ is the atmospheric density at an altitude of $0 m$ and is taken as 1.225$kg/m^3$ in this paper.

Considering the energy consumption of all track points, the energy consumption-related objective function can be expressed as follows.
\begin{equation}
	f_{\textup{fuel}} = \sum_{i=0}^{n-1} {fuel}_i.
\end{equation}
\subsubsection{Path height changes \textup{\cite{aitsaadiUAVPathPlanning2022a}}}
During a UAV mission, fewer UAV altitude changes mean fewer climb and descent phases, extending the life of the UAV power system. To a certain extent, this reflects the power cost. In this paper, we use the UAV altitude variation at all trajectory points as an objective function, as shown in (\ref{eqt:height_changes}).
\begin{equation}
	\label{eqt:height_changes}
	f_{\textup{height}} = \sum_{i=0}^{n-1} |z_{i+1}-z_{i}| .
\end{equation}

\subsubsection{Mission hover point distance}
In a UAV-assisted wireless sensor network, Ref. \cite{shen2022energy,liu2022drl,singh2022multi} presents the application of UAVs collecting urban sensor data, which requires specific UAV hover points (UHPs) for the missions so that the UAV collects sensor data while hovering. In this paper, we consider some preset UHPs, and the UAV needs to minimize the sum of the distances to all the preset UHPs when the UAV actually hovers. The UHP distance objective function can be expressed as follows.
\begin{equation}
	f_{\textup{distance}} = \sum_{k=0}^{K-1} \mathop{min}\limits_{i} ||\vec{p_i}-p_k^{job}||,
\end{equation}
where $p_k^{job}$ is the $k$th preset UHP and $\vec{p_i}$ is the $i$th discrete trajectory point. The closest point among the trajectory points to the preset UHP is taken as the actual hover point, and the sum of the distances of all preset UHPs and the corresponding actual hover points is taken as the objective function. Minimizing this value allows the UAV to complete the mission with higher quality. For example, when using a UAV to collect urban sensor data, it can be optimized to improve the completeness of the collected data; when using a UAV to build a temporary communication network, it can be optimized to improve the accuracy of the coverage area; and when using a UAV to take aerial photographs, it can be optimized to reduce the loss of imaging quality caused by positional deviation.

\subsection{Safety Decision-Maker}\label{safef}
Recently, UAV safety, such as uncontrollable third-party risks, has received much attention. For example, Pang\emph{ et al.} \cite{pangUAVPathOptimization2022a} and Hu\emph{ et al.} \cite{huRiskAssessmentModel2020} considered uncontrollable third-party risks brought by UAVs to people, vehicles, and buildings in urban areas, such as fatality risks, property damage, and noise pollution. Ghambari\emph{ et al.} \cite{ ghambariEnhancedNSGAIIMultiobjective2020} minimized the population density of the area through which the UAV passes to reduce potential third-party fatality risks, and Delamer\emph{ et al.} \cite{delamerSafePathPlanning2021} optimized the relationship between the flight trajectory and UAV positioning signals to improve the safety of UAV navigation during flight.

For the safety DM, the core demand is to ensure that the risk to urban residents and property from UAV operations is as small as possible. According to Ref. \cite{pangUAVPathOptimization2022a}, the main objectives of the safety decision are to minimize the risks to pedestrians, vehicles, and property and noise pollution.
\subsubsection{Fatality Risks \textup{\cite{pangUAVPathOptimization2022a}}}
Fatality risk refers mainly to the risk of endangering the safety of third parties, including the risk of pedestrian fatalities and the risk of vehicular fatalities. The fatality risk cost associated with pedestrian fatalities, i.e., $c_{r_p}$, is expressed as follows.
\begin{equation}\label{crp}
	c_{r_p}(x,y,z) = P_{crash}S_{h}\sigma_{p}(x,y)R^{P}_{f}	,
\end{equation}
where $P_{crash}$ mainly depends on the reliability of the UAV itself, including the reliability of hardware and software. $S_{h}$ is the size of the UAV crash impact area, and $\sigma_{p}(x,y)$ is the population density within the administrative unit, which is a location-dependent factor. $R^{P}_{f}$ is related to the kinetic energy of the impact and the obscuration factor. Ref. \cite{pangUAVPathOptimization2022a} shows the calculation in (\ref{eqt:fatal_risks}).
\begin{equation}
	\label{eqt:fatal_risks}
	R^{p}_{f}=\frac{1}{1+\sqrt{\frac{\alpha}{\beta}}(\frac{\beta}{\frac{1}{2}mv(z)^2})^{\frac{1}{4S_{c}}}},
\end{equation}
where $\alpha$ and $\beta$ are the energy with $S_{c}$ probability of death and the energy with $100\%$ of death, respectively, $m$ is the mass of the UAV, and $v$ is the velocity of the UAV when it hits the ground. The velocity $v$ when it hits the ground at an altitude of $z$ is obtained by (\ref{eqt:velocity}).
\begin{equation}
	\label{eqt:velocity}
	\begin{split}
		v(z)=\sqrt{\frac{2mg}{R_IS_{h}\rho(1-\exp^{\frac{zR_IS_{h}\rho}{m}})}},
	\end{split}
\end{equation}
where $m$ is the mass of the UAV and $g$ is the gravitational constant. $R_I$ is the drag coefficient related to the UAV type, and $\rho$ is the fluid density of air. It is easy to see that the population risk indicator is determined by the three-dimensional space in which the UAV is located. 

Similarly, the fatality risk index of vehicles can be expressed as $c_{r_v}(x,y,z) = P_{crash}S_{h}\sigma_{v}(x,y)R^{V}_{f}(z)$, which is based on the traffic density $\sigma_{v}(x,y)$ and the corresponding impact energy parameters of the vehicle $R^{V}_{f}(z)$ \cite{pangUAVPathOptimization2022a}. Finally, the two fatality risks are added to obtain the overall fatality risk, shown as follows.
\begin{equation}
	f_{\textup{fatal}} = \sum_{i=0}^{n}c_{r_p}(x_i,y_i,z_i)+\sum_{i=0}^{n}c_{r_v}(x_i,y_i,z_i).
\end{equation}
\subsubsection{Property Risk \textup{\cite{pangUAVPathOptimization2022a}}}
When the UAV is operating at high altitudes, it may collide with tall buildings in cities, especially those with high building density. The flight height is the main factor affecting the property risk. Existing work has shown that the building height distribution does not conform to the standard normal distribution but to the lognormal distribution because the building height is nonnegative and its distribution is not symmetrical. Based on these characteristics, the property risk index $c_{r_p,d}$ was established as follows.
\begin{equation}
	\psi(z_i;\mu,\sigma)=\frac{1}{z_i \sigma\sqrt{2\pi}}e^{-\frac{(\ln z_i-\mu)^2}{2\sigma^2}},
\end{equation}
\begin{equation}
	c_{r_p,d}(z_i)=\begin{cases}
		\psi(e^\mu) &{\text{if}}\ 0< z_i \le e^\mu \\
		\psi(z_i) &{\text{otherwise.}}
	\end{cases},
\end{equation}
\begin{equation}
	f_{\textup{eco}} = \sum_{i=0}^{n}c_{r_p,d}(z_i),
\end{equation}
where the function $\psi$ is a distribution function of the flight altitude. $f_{eco}$ is the final property risk objective value. $\mu$ and $\sigma$ are the lognormal distribution parameters of the building height.
\subsubsection{Noise Pollution \textup{\cite{pangUAVPathOptimization2022a}}}
Noise impact is an important third-party impact for city safety supervisors to consider. UAV operations can cause complaints when they significantly impact the productivity and livelihood of urban residents. As the flight altitude increases, the impact decreases below a threshold that does not affect people on the ground. The noise beyond a certain distance is considered not to cause noise pollution for third parties, and the noise pollution indicator can be expressed as the approximate value of spherical propagation, as follows.
\begin{equation}
	C_{noise}(x_i,y_i,z_i)=k\sigma_{p}(x_i,y_i)L_h\frac{1}{(z_i)^2+d^2},
\end{equation}
where $C_{noise}$ is the noise impact cost of UAV operations in a given airspace cell. $k$ is the conversion factor from the sound intensity to the sound level, $L_h$ is the noise produced by the UAV, and $\sigma_{p}(x_i,y_i)$ is the density of people at track point $(x_i,y_i,z_i)$. $d$ is the distance between the UAV and the area of interest; if the flight altitude exceeds a certain threshold, the noise is small, and its impact will not be included in the calculation of pollution. The noise threshold was set to 40 dB in the experiment \cite{torijaEffectsHoveringUnmanned2020}. Finally, the objective function of the noise impact can be expressed as follows.
\begin{equation}
	f_{\textup{noise}} = \sum_{i=0}^{n} C_{noise}(x_i,y_i,z_i).
\end{equation}
\subsection{Constraints}\label{con}
In this part, we mainly consider the path constraints due to the dynamic performance limitations of UAV flight. Therefore, these constraints must be satisfied for all DMs.
\begin{figure*}
	\centering  
	\includegraphics[scale=0.55,trim=0 0 0 0,clip]{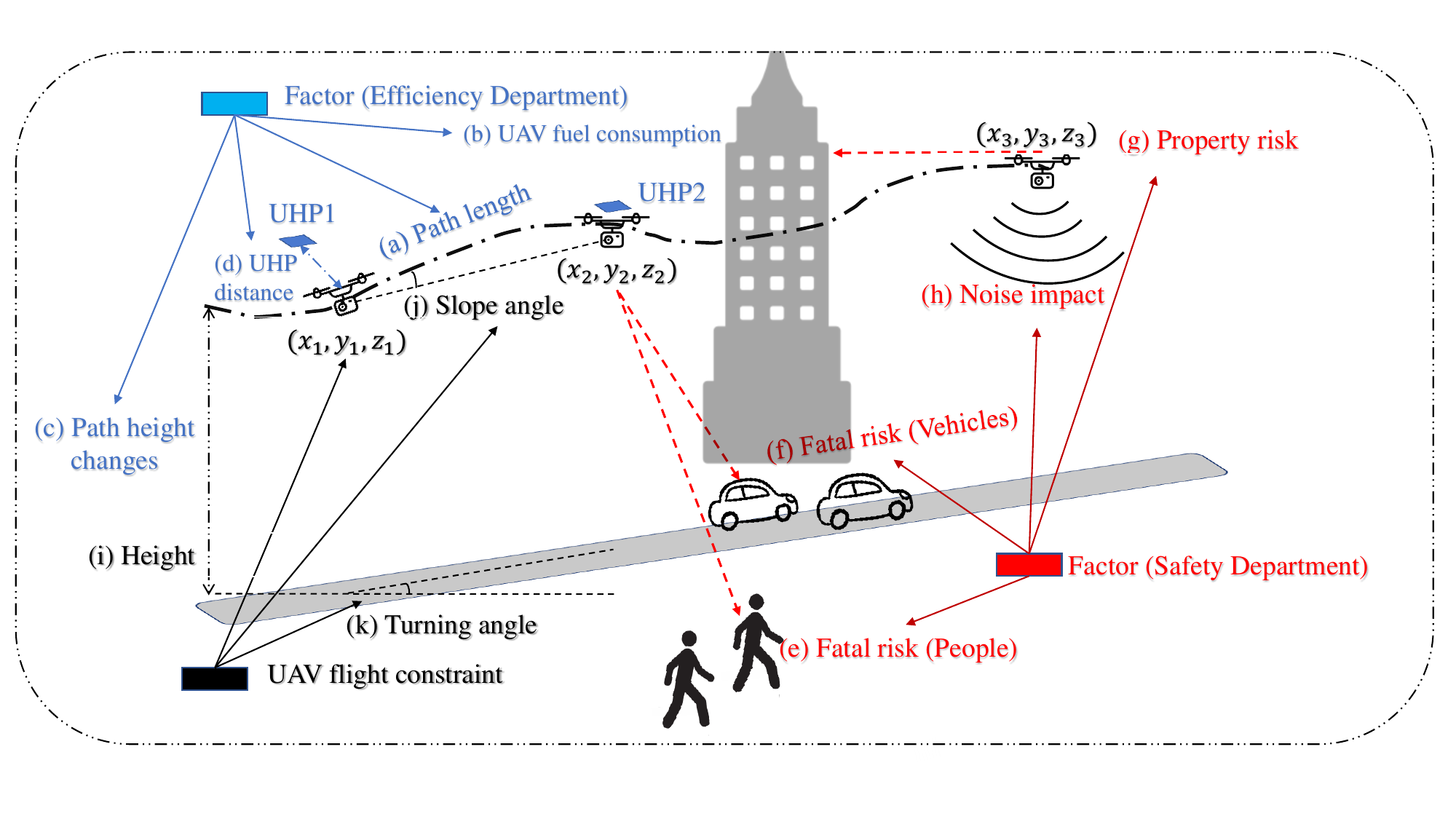}    
	\caption{The objectives and constraints of the UAV path planning problem from the perspective of the BPMOP. (a)-(d), which are marked in blue, are the objectives of the efficiency department. (e)-(h), which are marked in red, are the objectives of the safety department. (i)-(k), which are marked in black, are the constraints of the UAV.}
	\label{fig1}
\end{figure*}
\subsubsection{Terrain Constraints}
The UAV can only fly within a certain altitude range, as shown below.
\begin{equation}
	H_{min} \leq z_i \leq H_{max}.
\end{equation}
\subsubsection{Performance Constraints \textup{\cite{aitsaadiUAVPathPlanning2022a}}}
Let the projection of the $i$th track segment in the plane be $g_i'$; for a given track segment, the turning angle $\alpha_i$ between each track segment can be calculated as follows.
\begin{equation}
	\label{eqt:alpha}
	\alpha_i = \arccos{\frac{g_i'*g_{i+1}'}{||g_i'||*||g_{i+1}'||}}.
\end{equation}
Similarly, the slope angle formed by the $i$th and the $(i+1)$th segments can be defined as follows.
\begin{equation}
	\beta_i = \arctan{\frac{z_{i+1}-z_i}{||g_i'||}}.
\end{equation}

Due to the power performance of the UAV, the turning angle between the segments cannot be greater than the maximum turning angle, and the slope angle between the segments cannot be greater than the maximum slope angle; these are expressed as hard constraints, shown as follows.

\begin{equation}
	|\alpha_i| \leq \alpha_{max},|\beta_i| \leq \beta_{max},
\end{equation}
where $\alpha_i$ is the turning angle formed by the $i$th and $(i+1)$th segment, $\alpha_{max}$ is the maximum turning angle of the UAV, $\beta_i$ is the slope angle formed by the $i$th and $(i+1)$th segment, and $\beta_{max}$ is the maximum slope angle of the UAV.

\subsection{Model}
In the previous parts, we introduced the objectives of both efficiency and safety and the constraints related to the UAV performance limit, which will be used in the experiments of this paper. Combining these objectives into a MOP ignores the fact that there are two parties. In practical applications, the departments concerned with these two types of objectives are often relatively independent. The direct use of multiobjective models does not allow searching in the direction close to the optimal solutions of each DM, making the solution unacceptable to one of the departments. Therefore, this paper adopts BPMOPs to model the UAV path planning problem. To better test the algorithms, six application scenarios are considered, and six models are built. For convenience, Fig. \ref{fig1} shows the objectives and constraint factors for different decision-makers.

The efficiency objectives include the minimization of the flight path length $f_{\textup{length}}$, the minimization of the flight energy consumption $f_{\textup{fuel}}$, the minimization of the path height changes $f_{\textup{height}}$ and the minimization of the total sum of the mission hover point distance $f_{\textup{distance}}$. The safety objectives include the minimization of fatality risks $f_{\textup{fatal}}$, the minimization of property risk $f_{\textup{eco}}$ and the minimization of noise pollution $f_{\textup{noise}}$. Without loss of generality, the model for all six cases can be represented generically as follows.
\begin{equation}
	\label{eqt:case2}
	\begin{aligned}
		& \min (F_{eff},F_{safe}),\\
		& F_{eff}=(f_{11},f_{12},\dots),\\
		& F_{safe}=(f_{21},f_{22},\dots),\\
		&	S.t. \begin{cases}
			H_{min} \leq z_i \leq H_{max}, &  i = 1,\dots,n ;\\
			|\alpha_i| \leq \alpha_{max} , &  i = 1,\dots,n-1 ;\\
			|\beta_i | \leq \beta_{max} , &  i = 1,\dots,n-1 .\\
		\end{cases} \\
	\end{aligned}
\end{equation}

In the objectives of the efficiency DM, generally, minimizing the path length, minimizing the energy consumption, and minimizing the height changes belong to the same class of objective functions and do not have serious conflicts, but minimizing the total sum of mission hover point distances will sometimes cause more UAV energy consumption or more height changes. The efficiency DM selects one of the first three objectives as one objective and the minimization of the total sum of the mission hover point distance as the other objective to form a biobjective optimization problem.

In the objectives of the safety DM, flying UAVs at higher altitudes is good for minimizing noise impact and property risk but not for minimizing fatality risks. Generally, minimizing noise impact and property risk belong to the same class of objectives and conflict with minimizing fatality risk. Therefore, the DM chooses one of the first two objectives as one objective and minimizes fatality risk as the other objective to form a biobjective optimization problem. 

Considering that different DMs choose different objectives to form a biobjective optimization problem, the detailed objective function compositions are shown in Table \ref{table3}. Here, six biparty biobjective optimization problems are constructed for better testing of the algorithms. 

\begin{table}[H]
	\centering
	\caption{Case design}
	\begin{tabular}{ccc}
		\toprule[1.5pt]
		Problems &  Efficiency DM Objectives   & Safety DM Objectives        \\ 
		\cline{1-3}
		Case 1   &  $F_{eff}=(f_{\textup{length}},f_{\textup{distance}})$  & $F_{safe}=(f_{\textup{fatal}},f_{\textup{eco}})$   	\\
		Case 2   &  $F_{eff}=(f_{\textup{length}}+f_{\textup{height}},f_{\textup{distance}})$  & $F_{safe}=(f_{\textup{fatal}},f_{\textup{eco}})$      \\
		Case 3  &   $F_{eff}=(f_{\textup{fuel}},f_{\textup{distance}})$  & $F_{safe}=(f_{\textup{fatal}},f_{\textup{eco}})$     \\
		Case 4  &   $F_{eff}=(f_{\textup{length}},f_{\textup{distance}})$  & $F_{safe}=(f_{\textup{fatal}},f_{\textup{noise}})$      \\
		Case 5  &  $F_{eff}=(f_{\textup{length}}+f_{\textup{height}},f_{\textup{distance}})$ & 
		$F_{safe}=(f_{\textup{fatal}},f_{\textup{noise}})$ \\
		Case 6  &  $F_{eff}=(f_{\textup{fuel}},f_{\textup{distance}})$  & $F_{safe}=(f_{\textup{fatal}},f_{\textup{noise}})$  \\
		\bottomrule[1.5pt]
	\end{tabular}
	\label{table3}    
\end{table}
\section{Experimental Design}
\label{sec:design}
\subsection{Compared algorithms}
To solve MOPs with algorithms, the evolutionary algorithm (EA) framework has been widely used, such as NSGA-II \cite{debFastElitistMultiobjective2002a}. On the other hand, the clonal selection algorithm based on the immunity principle is a similar framework and includes algorithms such as NNIA, HEIA, and AIMA. These MOEAs use Pareto dominance relations to control convergence and use crowding distance information to control population diversity. In this paper, NSGA-II is used as a comparison algorithm in the experiments, while NNIA, HEIA, and AIMA are adapted for BPMOPs. We introduce them as follows.

\begin{enumerate}
	\item NSGA-II \cite{debFastElitistMultiobjective2002a}: This classical multiobjective evolutionary optimization algorithm uses nondominated sorting to guide the search direction in MOPs and combines the crowding distance information to maintain the diversity of the population.
	\item NNIA \cite{gongMultiobjectiveImmuneAlgorithm2008}: This classical multiobjective immune optimization algorithm, based on the framework of clonal selection algorithms, uses nondominated sorting to guide the search direction in MOPs and uses crowding distance information to allocate clonal resources to the population. The individuals with low density can be allocated more computational resources, thus maintaining the population diversity.
	\item HEIA \cite{linHybridEvolutionaryImmune2015}: This algorithm based on the NNIA, a crossover approach combining the differential evolution (DE) operator and the simulated binary crossover (SBX) operator, is used to 
make the algorithm search more widely
 when dealing with complex problems.
	\item AIMA \cite{lin2018adaptive}: This algorithm uses multiple DE strategies during the evolutionary process. The components of AIMA are designed following the framework of NNIA. AIMA adopts an adaptively selected DE strategy and the polynomial mutation operator.
\end{enumerate}

For BPMOPs, there are some algorithms based on the idea of finding the common Pareto optimal solutions of each decision-maker. Typically, OptMPNDS and OptMPNDS2 adopt different dominance hierarchy ranking methods to search for the optimal solutions in BPMOPs; they are used as comparison algorithms in experiments and described as follows.

\begin{enumerate}
	\item OptMPNDS \cite{liuEvolutionaryApproachMultiparty2020}: 
	This study is the first to solve multiparty multiobjective optimization problems. In detail, it is based on NSGA-II and adopts the numbers of nondominated layers of all DMs to determine the nondominated layers of the population in MPMOPs. The experimental results showed that OptMPNDS achieves better performance than NSGA-II in MPMOPs.
	\item OptMPNDS2 \cite{10.1007/978-3-030-78811-7_6}: 
	This algorithm directly regards the number of nondominated layers under different DMs as the objectives and adopts two rounds of nondominated ranking to calculate nondominated layers for solving MPMOPs, which yields lower time complexity and better experimental results.
	
\end{enumerate}	

At present, there is no algorithm that uses the clonal selection framework to solve BPMOPs. Unlike NSGA-II, most multiobjective immune algorithms (MOIAs) use the clonal selection mechanism to allocate the next-generation resources according to the crowding distance information. In this paper, we extend three typical multiobjective immune optimization algorithms, NNIA, HEIA, and AIMA, to multiparty multiobjective optimization algorithms, called BPNNIA, BPHEIA, and BPAIMA, respectively. The main processes of these three algorithms are the same as those of the original algorithms, with the difference that the nondominated sorting method in multiobjective optimization is replaced by biparty nondominated sorting in BPMOPs. The framework of these algorithms is given in Algorithm \ref{alg:BPMOEAs}. The main changes are shown below. 
\begin{algorithm}[]
	\caption{Framework}
	\label{alg:BPMOEAs}
	\textbf{Input:} $nC$ (clone size), $nA$ (activation size), $lb$, $ub$ (upper bound) \\
	\textbf{Output:} $MPS$ (the multiparty Pareto optimal solutions)\begin{algorithmic}[1]
		\State $P^0, f^0= \textbf{Initialization}(nC,lb,ub),t = 0$;
		\State $F^0 = \textbf{MPNDS2}(P^0, f^0)$;
		\While {$FE$ is not reached}
		\State $\mathcal{A}, \mathcal{AF} = \textbf{Activate}(P^t, nA, F^t)$;
		\State $\mathcal{C} = \textbf{Clone}(\mathcal{A}, \mathcal{AF}, nC)$;
		\State $\mathcal{O} = \textbf{Cross\&Mutate}(\mathcal{A}, \mathcal{C})$;
		\State $f_O = \textbf{Eval}(O)$;$P^t = P^t \cup O$;$f^{t+1} = f^t \cup f^O$;
		\State $F^{t+1} = \textbf{MPNDS2}(P^t, f^t)$;
		\State $P^{t+1},f^{t+1} = \textbf{Selection}(P^{t+1},F^{t+1},f^{t+1},nC)$;
		\State $t = t+1$;
		\EndWhile 
		\State set $MPS$ as the solutions that are Pareto optimal for all DMs in $P^{t+1}$;
	\end{algorithmic}
\end{algorithm}

\textbf{Initialization}: This operation uses uniformly distributed random numbers to generate individuals in the space composed of the upper and lower bounds of the decision variables and evaluates these individuals. This operation is consistent with the operation in the original algorithms and can be formulated as follows \cite{gongMultiobjectiveImmuneAlgorithm2008,linHybridEvolutionaryImmune2015,linAdaptiveImmuneinspiredMultiobjective2018}.

\begin{equation}
	P^0_{i,j} = lb_j+(ub_j-lb_j)*rand,
\end{equation}
where $P^0$ is the initialized population, $ub$ is the upper bound of the decision variable, and $lb$ is the lower bound of the decision variable. The subscript ${i,j}$ represents the $j$-th decision variable of the $i$-th individual. 

\textbf{MPNDS2} \cite{10.1007/978-3-030-78811-7_6}: The sort operation, which regards the number of nondominated layers under different DMs as the objectives, adopts two rounds of nondominated ranking to calculate nondominated layers for solving MPMOPs. The pseudocode of MPNDS2 is given in Algorithm \ref{alg:MPNDSort}.
\begin{algorithm}[H]
	\caption{MPNDS2 \cite{10.1007/978-3-030-78811-7_6}}
	\label{alg:MPNDSort}
	\textbf{Input:} $P_t, F_t=(F_{safe}(x),F_{eff}(x))$ \\
	\textbf{Output:} $\mathcal{F}_{t+1}$ (the true number of layers in the BPMOP)
	\begin{algorithmic}[1]
		\State $\mathcal{L}_{eff} = \text{NonDominatedSorting}(P_t,F_{eff})$;
		\State $\mathcal{L}_{safe} = \text{NonDominatedSorting}(P_t,F_{safe})$;
		\State $\mathcal{F}_{t+1} = \text{NonDominatedSorting}(P_t,(\mathcal{L}_{safe},\mathcal{L}_{eff}))$;
	\end{algorithmic}
\end{algorithm}

\textbf{Activate}: The activate operation uses 
the front part of the individuals sorted by MPNDS2 as the activated individuals. The activate operation is used 
in multiparty MOIAs. The sorting results of MPNDS2 are used instead of the sorting results from fast nondominated sorting in NNIA \cite{gongMultiobjectiveImmuneAlgorithm2008}, HEIA \cite{linHybridEvolutionaryImmune2015}, and AIMA \cite{lin2018adaptive}. Unlike these ordinary MOIAs for MOPs, which only activate the first nondominated layer of individuals, in MPMOPs, the first nondominated layer of individuals sorted by MPNDS2 usually only makes up a small portion of the population. Therefore, in MPNNIA, MPHEIA, and MPAIAM, all individuals with a sorting rank less than $nA$ are activated.
\begin{equation}
	A=P^{sorted}(1:nA,:),
\end{equation}
where 
$P^{sorted}$ is the set of individuals 
of population P sorted by MPNDS2($\cdot$) and $nA$ is the preset number of activated individuals.


\textbf{Clone}: The allocation of clonal individuals is determined by multiple nondominated sorting layers and crowding distance information. Unlike previous algorithms, such as NNIA \cite{gongMultiobjectiveImmuneAlgorithm2008}, HEIA \cite{linHybridEvolutionaryImmune2015}, and AIMA \cite{lin2018adaptive}, which only use the crowding distance information, to increase the convergence pressure, the number of layers is also included as an indicator of resource allocation in the calculation of the number of clones. The formula for the convergence indicator is as follows.
\begin{equation}
	p_i = \max_{j = 1,\dots,nA}(AF_j)-AF_{i},
\end{equation}
where $p_i$ is the convergence indicator of the $i-th$ activated individuals. $AF_{i}$ indicates the multiple nondominated sorting layers of the $i-th$ activated individuals. The individuals with larger convergence indicators have better convergence. Based on the convergence indicator, the formula for the number of clones per individual is as follows.
\begin{equation}
	\label{eq:ri}
	Nums_i = \lceil nC*(\frac{\textup{cd}(a_i)+ p_i}{\sum_{nA}^{j} (\textup{cd}(a_j) + p_j)}) \rceil ,
\end{equation}
where $Nums(i)$ is the number of clones of the $i-th$ activated individuals. Function $\textup{cd}()$ is the function that calculates the crowding distance, and $\textup{cd}(a_i)$ is the crowding distance of the $i-th$ activated individuals. $p_i$ is the convergence indicator of the $i-th$ activated individuals.

\textbf{Cross\&Mutate}: For crossover, BPNNIA uses the simulated binary crossover (SBX) operator, BPHEIA uses the SBX operator or the differential evolution operator (DE2) with a certain probability, and BPAIMA uses adaptive multiple differential evolution operators (DE1, DE2 and DE3). In terms of mutation, all of them use the polynomial mutation operator (PM). More details can be found in the papers \cite{gongMultiobjectiveImmuneAlgorithm2008,linHybridEvolutionaryImmune2015,lin2018adaptive}.

\begin{footnotesize}
	\begin{equation}
		DE1:O_{i,j}=\begin{cases}
			C_{i,j}+F_1*(A^1_{i,j}-A^2_{i,j})
			\\ + F_1*(A^3_{i,j}-A^4_{i,j}) &{\text{if}}\ rand<CR_1\ or\ j==j_{r} \\
			C_{i,j} &{\text{otherwise.}}
		\end{cases},
	\end{equation}
\end{footnotesize}
\begin{footnotesize}
	\begin{equation}
		DE2:O_{i,j}=\begin{cases}
			C_{i,j}+F_2*(A^1_{i,j}-A^2_{i,j}) &{\text{if}}\ rand<CR_2\ or\ j==j_{r} \\
			C_{i,j} &{\text{otherwise.}}
		\end{cases},
	\end{equation}
\end{footnotesize}
\begin{footnotesize}
	\begin{equation}
		DE3:O_{i,j}=\begin{cases}
			C_{i,j}+F_2*(A^1_{i,j}-A^2_{i,j}) &{\text{if}}\ rand<CR_3\ or\ j==j_{r} \\
			C_{i,j} &{\text{otherwise.}}
		\end{cases},
	\end{equation}
\end{footnotesize}

\begin{footnotesize}
	\begin{equation}
		SBX:O_{i,j}=\begin{cases}
			0.5*[(1+\delta)*C_{i,j}+(1-\delta)*A^1_{i,j}] &{\text{if}}\ rand<P_c \\
			C_{i,j} &{\text{otherwise.}}
		\end{cases},
	\end{equation}
\end{footnotesize}
\begin{equation}
	\delta=\begin{cases}
		(rand * 2)^{\frac{1}{1+disC}} &{\text{if}}\ rand<0.5 \\
		(1 / (2-rand*2))^{\frac{1}{1+disC}} &{\text{otherwise.}}
	\end{cases},
\end{equation}
where $O$ represents the individual generated after the operation, $C$ represents the cloned individuals, and $A^1$-$A^4$ represents the different activated individuals. The subscript ${i,j}$ represents the $j$-th decision variable of the $i$-th individual or corresponding offspring. $CR_1$ - $CR_3$ indicate the crossover rates, which are 0.9, 0.5 and 0.1. $F_1$ - $F_3$ are the scaling factors, which are 0.7, 0.5 and 0.1. $P_c$ is the crossover rate, which is the reciprocal of the decision variables. 
$\delta$ is calculated from the cross-distance $disM$, as shown in the following formula, where $disC$ is the distance of the crossing.

\textbf{Selection}: In the operation of retaining populations, we first use MPNDS2 for sorting. Then, we add 
the individuals
 to the next generation's populations layer by layer according to the number of sorted layers until the size exceeds the number of individuals that can be preserved. At this point, we use the crowding distance to select some of the individuals in this layer to be added to the next generation's populations and discard the individuals whose crowding distances are too small. In other words, the crowding distance information is used only in the last retained layer. The selection operation in MOIAs for multiparty cases is similar to the selection operation in regular MOIAs, except that the sorting results from MPNDS2 are used instead of fast nondominated sorting as in NNIA \cite{gongMultiobjectiveImmuneAlgorithm2008}, HEIA \cite{linHybridEvolutionaryImmune2015}, and AIMA \cite{lin2018adaptive}.

\subsection{Metric}
For MPMOPs, Liu\emph{ et al.} \cite{liuEvolutionaryApproachMultiparty2020} used the inverted generational distance (IGD \cite{liuEvolutionaryApproachMultiparty2020}) to measure the performance of the algorithms. However, for real-world problems, there is often no definite true PF to calculate the IGD metric. To obtain the performance of different algorithms on BPMOPs, in this paper, the meanHV metric is used, which is based on the hypervolume (HV) metric \cite{zitzler1999multiobjective}.

For general MOPs, the HV metric is defined as the supervolume constituted by the normalized solution set. Suppose $HV_i$ is the HV metric of the solution set on the set of objectives of the $i$th DM; then, \textit{meanHV} is defined as the mean of $HV_i$ of the solution set for all decision-makers, as follows.
\begin{equation}
\textit{meanHV} = \frac{\sum_{i}^K HV_i}{K}.
\end{equation}

The \textit{meanHV} metric examines the performance of the solution set for different decision-makers, and the mean performance is chosen as the performance metric of the solution set. 

\section{Experiments}	
\label{sec:ex}
\subsection{Experimental Setup}
\begin{figure}[]
\centering  
\includegraphics[scale=0.6,trim=110 270 0 240,clip]{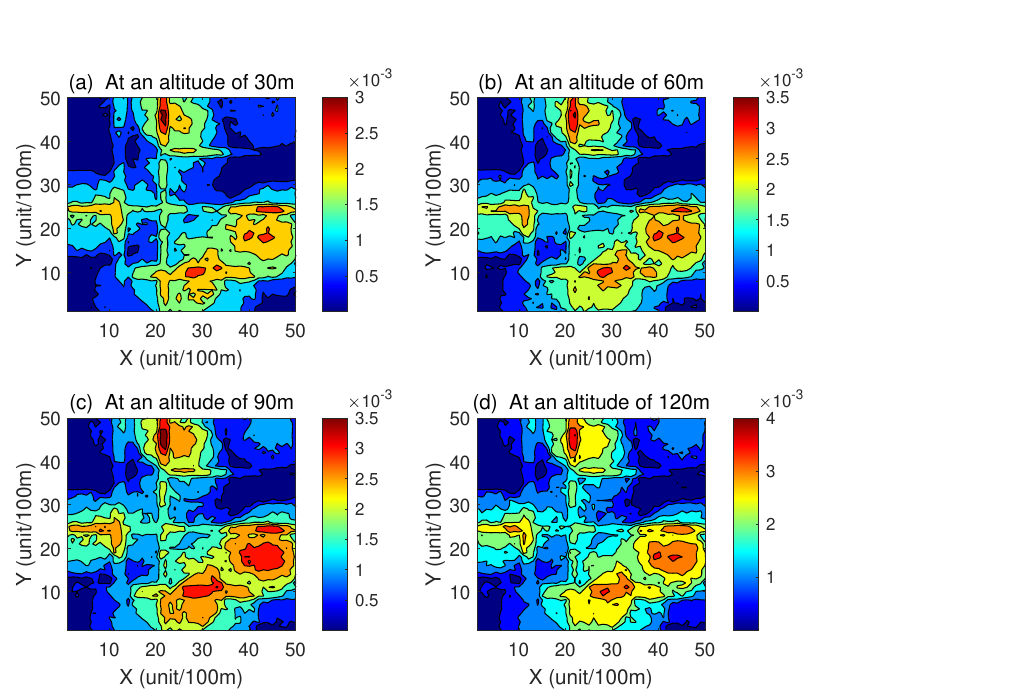}    
\caption{Third-party fatality risk distribution maps}
\label{fig2}
\end{figure}

\paragraph{Case Data} In terms of building data generation, this paper uses a lognormal distribution to describe the distribution of building heights. The parameter of lognormal distribution $\mu$ is 3.04670, and $\sigma$ is 0.76023 \cite{pangUAVPathOptimization2022a}. The distribution of the population is related to the core metropolitan area, in which the population is concentrated. Pang\emph{ et al.} \cite{pangUAVPathOptimization2022a} used a radial basis model to simulate population density, and the same approach is used here to generate the experimental data. The fatality risk distribution map data are visualized in Fig. \ref{fig2}.
\begin{figure}[]
\centering  
\includegraphics[scale=0.54,trim=13 0 0 30,clip]{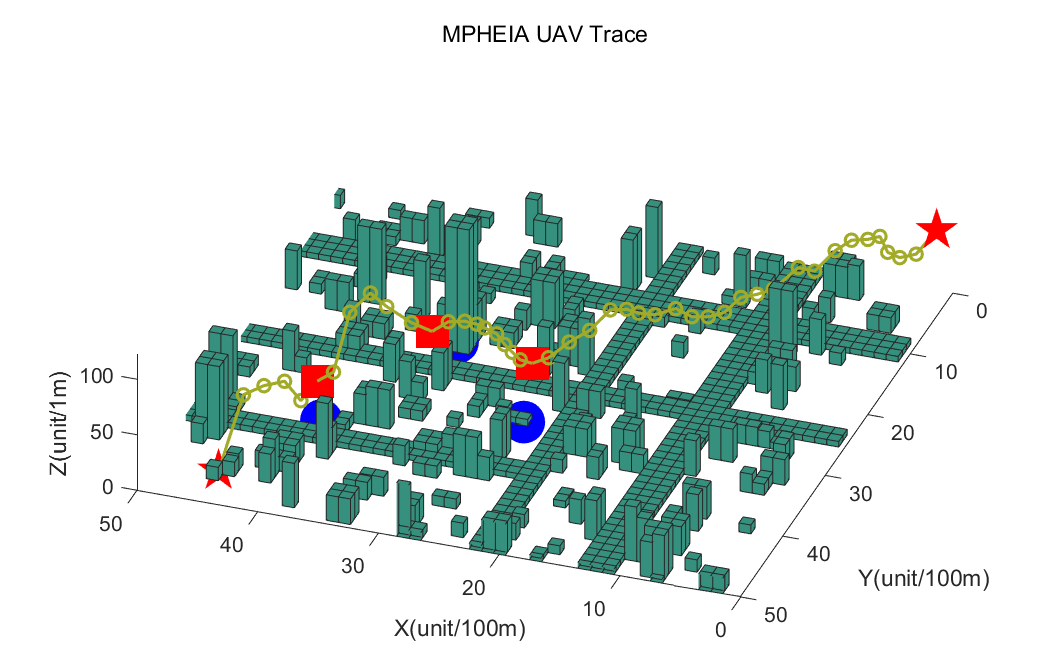}    
\caption{Building, roads, UHPs, and one of the BPAIMA solutions}
\label{fig3}
\end{figure}

\paragraph{Model Parameters}
The parameters related to the calculation of UAV fuel consumption are set to the same values as in Ref. \cite{ghambariEnhancedNSGAIIMultiobjective2020}, and the parameters related to third-party risks are set to the same values as in Ref. \cite{pangUAVPathOptimization2022a}. The other parameters are listed as follows. The maximum turning angle $\alpha_{max}$ is $\pi/3$, the maximum slope angle $\beta_{max}$ is $\pi/4$, the atmospheric density $\rho_{A}$ is $1.225 kg/m^3$, the flight speed $v$ is $10 m/s $, the rotating slope area $S_b$ is $0.1 m^2$, the number of rotating paddles $n$ is 4, and the UAV quality $m$ is $1.38 kg$ \cite{pangUAVPathOptimization2022a}. The starting point of the mission is (1,1), and the endpoint is (49, 49). The preset UHPs are (25, 30), (34, 20), and (40, 35).

\paragraph{Algorithm Parameters}
In the experimental parameter settings of the algorithms, some of the parameters of NSGA-II and the MPMOEAs mainly use the default settings in the PlatEMO platform \cite{tianPlatEMOMATLABPlatform2017} and original references \cite{liuEvolutionaryApproachMultiparty2020,10.1007/978-3-030-78811-7_6}. NSGA-II, OptMPNDS, OptMPNDS2, BPNNIA, and BPHEIA use the simulated binary crossover operator, the distribution index of the simulated binary crossover operator $disC$ is 20, and the crossover probability $P_c$ is 1. These algorithms also use the polynomial mutation operator; the distribution index of the polynomial mutation operator $disM$ is 20, and the mutation probability $P_m$ is the reciprocal of the number of decision variables ($1/88$ in Cases 1-6). In addition, BPHEIA uses the rand/1/bin DE operator ($CR = 0.5$, $F = 0.5$). BPAIMA uses the rand/2/bin DE operator ($CR = 0.9$, $F = 0.7$), rand/1/bin DE operator ($CR = 0.5$, $F = 0.5$), and rand/1/bin DE operator ($CR = 0.1$, $F = 0.5$). For the immune algorithms BPNNIA, BPHEIA and BPAIMA, the number of activated antibodies $A$ is 20. For all algorithms, the population size is 105, and the number of evaluations is 80000. Each experiment is run independently 30 times.

The code of this study is available at \url{https://github.com/MiLab-HITSZ/2023ChenMPUAV}.
\subsection{Experimental results and analysis}
\subsubsection{Metric-based analysis}
Table \ref{table2} shows the final experimental results of all algorithms in Cases 1-6, where each cell indicates the mean value and variance of the \textit{meanHV} of the experimental results. The results of the algorithm with the best performance in each case are shown in bold. For the \textit{meanHV} index, the larger the mean value is, the better the convergence and diversity of the algorithm's solutions. In this experiment, the solution results of the biparty multiobjective optimization algorithms are significantly better than those of NSGA-II. Additionally, the biparty multiobjective immune optimization algorithm BPAIMA has better performance than other algorithms. That is, BPAIMA performs best in Cases 1-6.
\begin{table*}[]
\centering
\caption{\textit{meanHV} metrics of different algorithms}
\begin{tabular}{c|c|c|c|c|c|c}
	\toprule[1.5pt]
	\multirow{2}{*}{Problems}    & \multicolumn{6}{c}{Algorithms} \\
	\cline{2-7}
	\multirow{2}{*}{}    &  NSGA-II  & OptMPNDS & OptMPNDS2 & BPNNIA & BPHEIA & BPAIMA     \\  
	\hline
	Case 1 & 0.0981($\pm$0.0016)  &  0.1531($\pm$0.0012) &   0.1670($\pm$0.0013) &  0.1665($\pm$0.0010)   & 0.1685($\pm$0.0012) & \textbf{0.1852($\pm$0.0014)} \\
	Case 2 &0.1073($\pm$0.0022)&    0.1686($\pm$0.0011) &   0.1787($\pm$0.0013)&   0.1807($\pm$0.0019) &   0.1761($\pm$0.0015) & \textbf{0.2102($\pm$0.0014)} \\
	Case 3 &0.0935($\pm$0.0012)&    0.1544($\pm$0.0011) &   0.1593($\pm$0.0009) &  0.1697($\pm$0.0010)&    0.1663($\pm$0.0009) & \textbf{0.1959($\pm$0.0009)}\\
	Case 4 &0.1163($\pm$0.0022)&    0.1882($\pm$0.0008)  &  0.1900($\pm$0.0009 ) &  0.1938($\pm$0.0013) &   0.1816($\pm$0.0008 ) & \textbf{0.1979($\pm$0.0015)}\\
	Case 5 &0.0817($\pm$0.0012)&   0.1778($\pm$0.0018)  &  0.1857($\pm$0.0015)&   0.1855($\pm$0.0011)  &  0.1796($\pm$0.0013) & \textbf{0.1924($\pm$0.0020)}\\
	Case 6 &0.0865($\pm$0.0012)&   0.1734($\pm$0.0005) &   0.1841($\pm$0.0004) &   0.1839($\pm$0.0004) &   0.1796($\pm$0.0004) & \textbf{0.1908($\pm$0.0005)}\\ 
	\bottomrule[1.5pt]
\end{tabular}
\label{table2}    
\end{table*}
\begin{figure*}[]
\centering  
\includegraphics[scale=0.56,trim=100 0 0 0,clip]{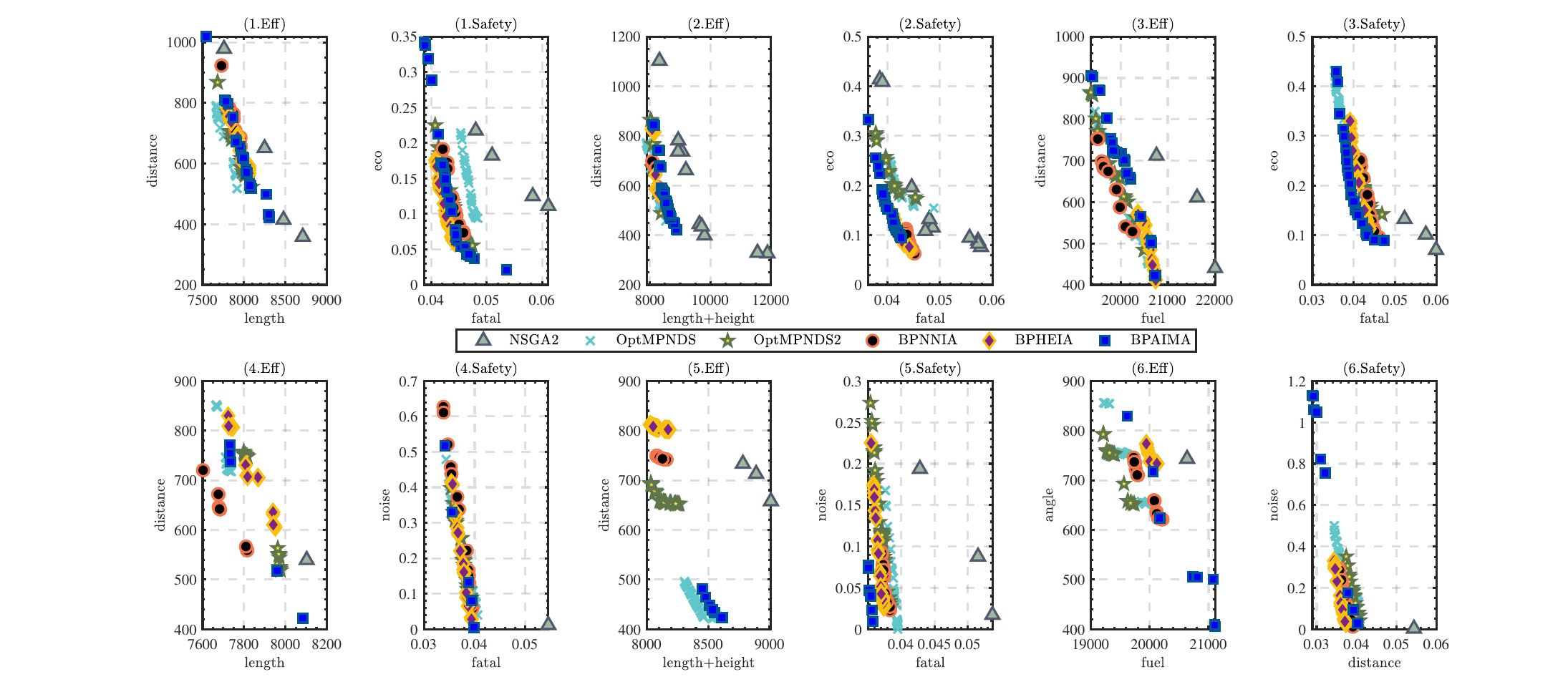}    
\caption{Optimal solutions obtained by MOEAs or BPMOEAs in each case}
\label{fig4}
\end{figure*}

Fig. \ref{fig3} shows a solution obtained by the BPAIMA algorithm, which avoids the high-risk area, has a continuous trajectory, and performs reasonably well. Fig. \ref{fig4} shows the results of solving Cases 1-6 with the ordinary multiobjective optimization algorithm NSGA-II; existing multiparty multiobjective optimization algorithms OptMPNDS and OptMPNDS2; and biparty multiobjective algorithms BPNNIA, BPHEIA and BPAIMA, which are proposed in this paper. For each algorithm, the optimal solutions are eventually obtained from among the final population based on the idea that the acceptable solutions are not Pareto dominated from the perspective of at least one DM. 

From the results, it is intuitively clear that using NSGA-II to solve the problem directly is unsatisfactory to one or two DMs. This result is mainly because NSGA-II does not have a mechanism to control the population convergence for different DMs in the solution process. For biparty multiobjective algorithms, the solution results have better convergence, with those of BPAIMA being the closest to the frontier of the decision-makers in Cases 1-6. The visualization shown in Fig. \ref{fig4} is consistent with the results of the meanHV metric shown in Table \ref{table2}.
\subsubsection{Convergence and process analysis}
From the meanHV of the result, it is seen that the biparty multiobjective optimization algorithms are better than the ordinary multiobjective optimization algorithms. Furthermore, some visualization results based on the convergence process and the final solution set are given to verify this conclusion more intuitively. 
\begin{figure}[h]
	\centering  
	\includegraphics[scale=0.60,trim=80 220 0 220,clip]{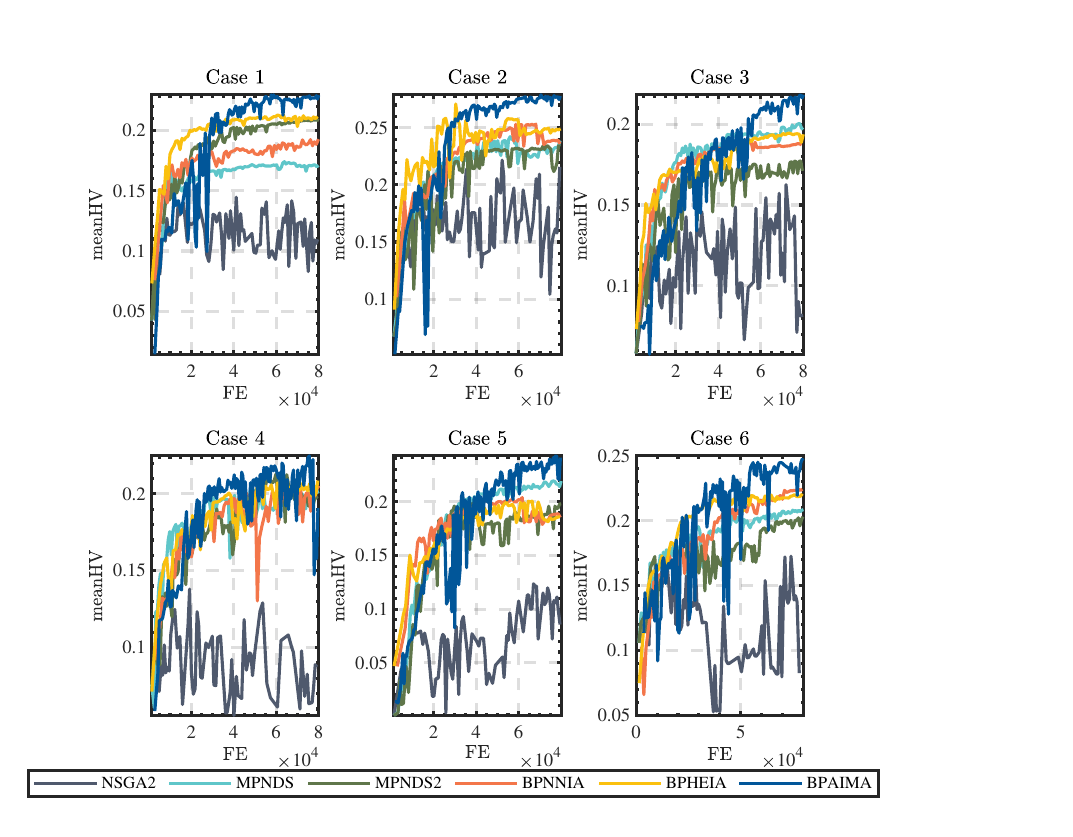}    
	\caption{Convergence performance of different algorithms on \textit{meanHV}}
	\label{convergence}
\end{figure}

Figure \ref{convergence} shows the meanHV of different algorithms during the evolution process in solving different cases, where NSGA-II has strong fluctuations in the meanHV of each generation. Because NSGA-II cannot correctly retain better individuals in BPMOPs, the meanHV did not increase effectively but fluctuated considerably. In contrast, for multiparty multiobjective evolutionary algorithms, the algorithms could guarantee final convergence, and the BPAIMA algorithms had the best performance.
\begin{figure}[h]
	\centering  
	\includegraphics[scale=0.45,trim=0 13 0 13,clip]{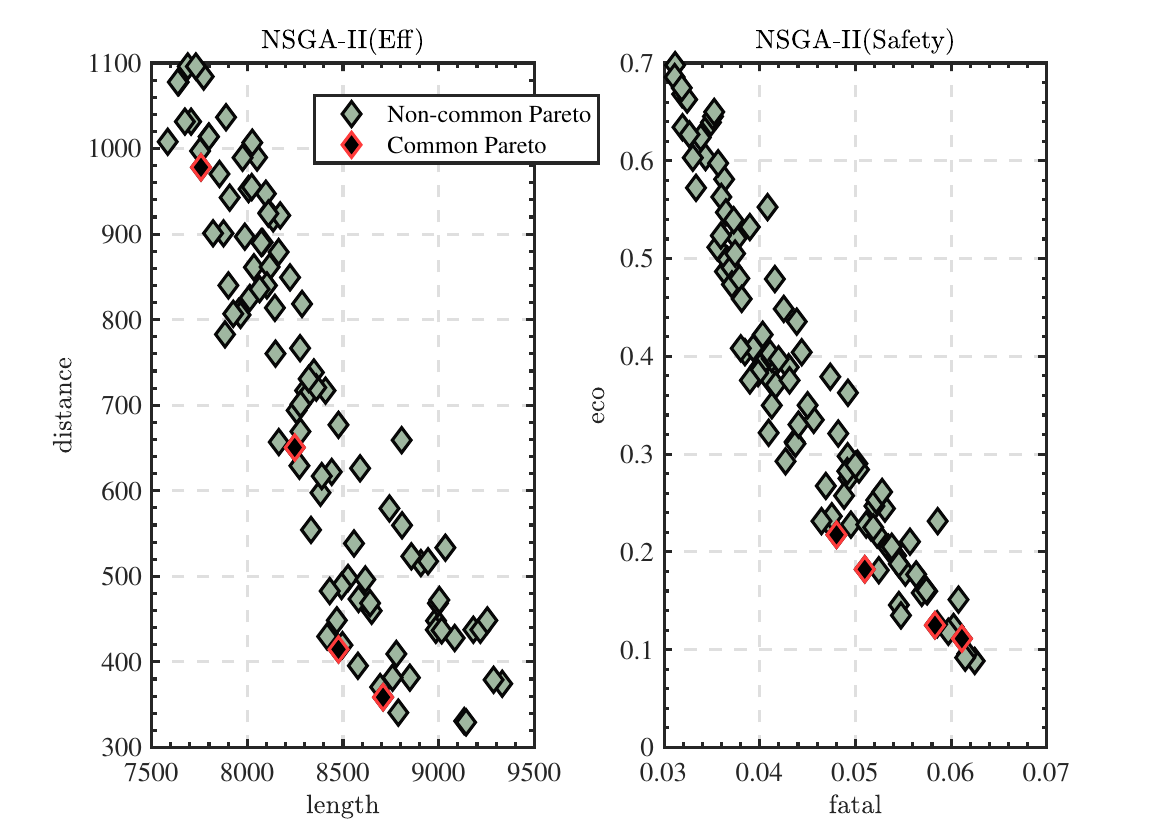}    
	\caption{Final population of NSGA-II in solving case 1}
	\label{NSGACASE1}
\end{figure}
\begin{figure}[h]
	\centering  
	\includegraphics[scale=0.48,trim=0 220 0 220,clip]{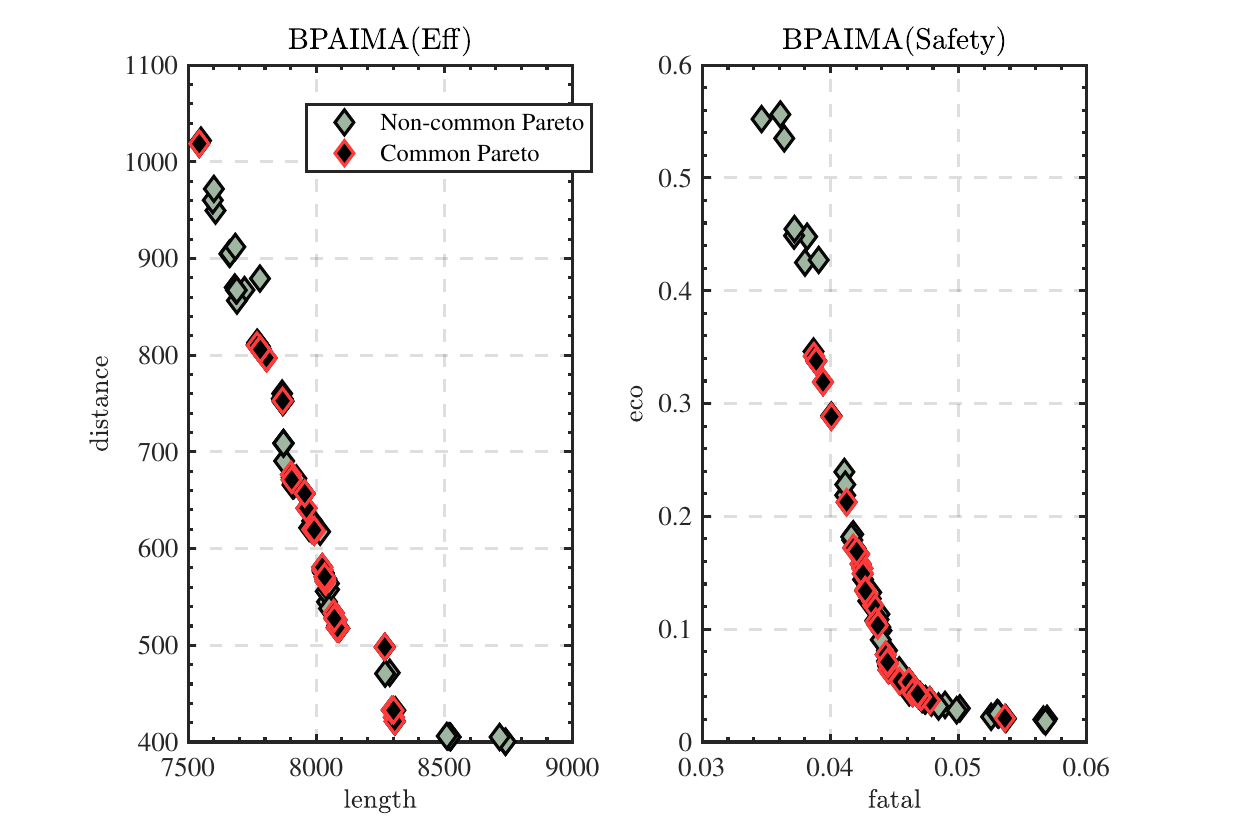}    
	\caption{Final population of BPAIMA in solving case 1}
	\label{BPAIMACASE1}
\end{figure}

Not only does the evolution of the population during the process demonstrate the shortcomings of ordinary multiobjective optimization algorithms, but the final solution set also demonstrates the same phenomenon. Fig. \ref{NSGACASE1} shows the final solution set of Case 1 solved by NSGA-II, and Fig. \ref{BPAIMACASE1} shows the final solution set of Case 1 solved by BPAIMA. The ordinary multiobjective optimization algorithm does not have a preference for the common Pareto front, which accounts for only a small fraction of all solutions. In contrast, the biparty multiobjective algorithms can obtain more good solutions for both DMs.

\subsection{Complexity Analysis}
The time complexity of the MPMOIAs comes from two main components. The first part comes from the time cost incurred during individual evaluations. In each iteration of the immune algorithm, the number of evaluations per iteration may vary. However, during the total process of the algorithm, the number of evaluations is equal to the preset parameter of the maximum number of evaluations. The time cost of this part is the same as that of the other algorithms.

The second component is the time consumption generated by the nondominated sorting of the algorithm. This part mainly concerns the additional time cost incurred when using MPNDS2($\cdot$) to replace the fast nondominated sorting in NSGA-II \cite{debFastElitistMultiobjective2002a}.

The time complexity of the nondominated sorting operation of NSGA-II is $\mathcal{O}(mn^2)$, where $m$ is the number of objectives and $n$ is the population size. For the multiparty multiobjective optimization problem with $m$ optimization objectives and $K$ decision-makers, operation MPNDS2($\cdot$) uses $K+1$ fast nondominated sorting operations of NSGA-II. In detail, the complexity of $K$ fast nondominated sorting operations with $\frac{m}{k}$ objectives is $\mathcal{O}(K\frac{m}{K}n^2)$, which is equivalent to $\mathcal{O}(mn^2)$. The complexity of one fast nondominated sorting operation with $K$ objectives is $\mathcal{O}(Kn^2)$. Therefore, the time complexity of MPNDS2($\cdot$) is $\mathcal{O}((K+m)n^2)$.

\begin{figure}[]
	\centering  
	\includegraphics[scale=0.6,trim=100 295 100 295,clip]{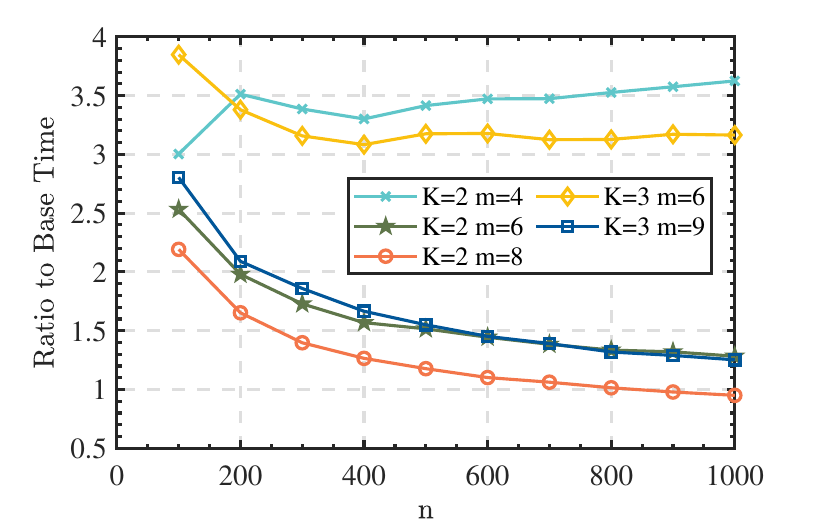}    
	\caption{Actual time expenditure ratios in different situations}
	\label{time}
\end{figure}

Because the number of decision-makers $K$ is often smaller than the number of objectives $m$, this conclusion can be further simplified. The complexity of operation MPNDS2($\cdot$) is $\mathcal{O}(mn^2)$ when $K\leq m$, which is the same as that of the fast nondominated sorting operation of NSGA-II. Fig. \ref{time} shows the ratio of the actual time cost of MPNDS2($\cdot$) to the constant scale of the base time (the actual time cost of the nondominated sorting operation of NSGA-II) with different numbers of DMs $K$, different numbers of objectives $m$, and different population sizes $n$.

Therefore, the theoretical time complexities of MPNNIA, MPHEIA, and MPAIMA are the same as those of NNIA, HEIA, and AIMA.

\section{Conclusion}
\label{sec:con}
This paper provides a novel biparty multiobjective perspective for UAV path planning problems. Specifically, a biparty multiobjective UAV path planning (BPMO-UAVPP) model is set up in this paper. This study is the first to embed a biparty multiobjective optimization model into UAV path planning. In addition, three optimization algorithms, i.e., BPNNIA, BPHEIA, and BPAIMA, are designed to solve BPMO-UAVPP problems, which are adapted from NNIA, HEIA, and AIMA, respectively. The experimental results show that BPAIMA has better performance in solving biparty multiobjective UAV path planning. In the future, we will deeply explore this problem and propose more efficient optimization algorithms to solve the problem.

\bibliographystyle{IEEEtran}
\bibliography{ref}

@article{lin2018adaptive,
  title={An adaptive immune-inspired multi-objective algorithm with multiple differential evolution strategies},
  author={Lin, Qiuzhen and Ma, Yueping and Chen, Jianyong and Zhu, Qingling and Coello, Carlos A Coello and Wong, Ka-Chun and Chen, Fei},
  journal={Information Sciences},
  volume={430},
  pages={46--64},
  year={2018},
  publisher={Elsevier}
}

@article{singh2022multi,
  title={Multi-objective {NSGA-II} optimization framework for {UAV} path planning in an {UAV}-assisted {WSN}},
  author={Singh, Manish Kumar and Choudhary, Amit and Gulia, Sandeep and Verma, Anurag},
  journal={The Journal of Supercomputing},
  pages={1--35},
  year={2022},
  publisher={Springer}
}

@article{shen2022energy,
  title={Energy-Aware Dynamic Trajectory Planning for {UAV}-Enabled Data Collection in {mMTC} Networks},
  author={Shen, Lingfeng and Wang, Ning and Zhang, Di and Chen, Jun and Mu, Xiaomin and Wong, Kon Max},
  journal={IEEE Transactions on Green Communications and Networking},
  volume={6},
  number={4},
  pages={1957--1971},
  year={2022},
  publisher={IEEE}
}

@article{liu2022drl,
  title={{DRL-UTPS: DRL}-based Trajectory Planning for Unmanned Aerial Vehicles for Data Collection in Dynamic {IoT} Network},
  author={Liu, Run and Qu, Zhenzhe and Huang, Guosheng and Dong, Mianxiong and Wang, Tian and Zhang, Shaobo and Liu, Anfeng},
  journal={IEEE Transactions on Intelligent Vehicles},
  year={2022},
  publisher={IEEE}
}

@inproceedings{song2022multiobjective,
  title={On Multiobjective Knapsack Problems with Multiple Decision Makers},
  author={Song, Zhen and Luo, Wenjian and Lin, Xin and She, Zeneng and Zhang, Qingfu},
  booktitle={Proceeding of 2022 IEEE Symposium Series on Computational Intelligence (SSCI)},
  pages={156--163},
  year={2022},
  organization={IEEE}
}

@inproceedings{10.1007/978-3-030-78811-7_6,
  title = {A New Evolutionary Approach to Multiparty Multiobjective Optimization},
  booktitle = {Proceeding of Advances in Swarm Intelligence},
  author = {She, Zeneng and Luo, Wenjian and Chang, Yatong and Lin, Xin and Tan, Ying},
  editor = {Tan, Ying and Shi, Yuhui},
  year = {2021},
  pages = {58--69},
  publisher = {Springer International Publishing},
  address = {Cham},
  isbn = {978-3-030-78811-7}
}

@article{aitsaadiUAVPathPlanning2022a,
  title = {{UAV} Path Planning Using Optimization Approaches: A Survey},
  author = {Ait Saadi, Amylia and Soukane, Assia and Meraihi, Yassine and Benmessaoud Gabis, Asma and Mirjalili, Seyedali and {Ramdane-Cherif}, Amar},
  year = {2022},
  month = apr,
  journal = {Archives of Computational Methods in Engineering},
  issn = {1134-3060, 1886-1784},
  doi = {10.1007/s11831-022-09742-7}
}

@article{besada-portasEvolutionaryTrajectoryPlanner2010,
  title = {Evolutionary Trajectory Planner for Multiple {UAVs} in Realistic Scenarios},
  author = {{Besada-Portas}, E and {de la Torre}, L and {de la Cruz}, J M and {de Andr{\'e}s-Toro}, B},
  year = {2010},
  month = aug,
  journal = {IEEE Transactions on Robotics},
  volume = {26},
  number = {4},
  pages = {619--634},
  issn = {1552-3098, 1941-0468},
  doi = {10.1109/TRO.2010.2048610}
}

@article{debFastElitistMultiobjective2002a,
  title = {A Fast and Elitist Multiobjective Genetic Algorithm: {NSGA-II}},
  author = {Deb, K. and Pratap, A. and Agarwal, S. and Meyarivan, T.},
  year = {2002},
  month = apr,
  journal = {IEEE Transactions on Evolutionary Computation},
  volume = {6},
  number = {2},
  pages = {182--197},
  issn = {1089778X},
  doi = {10.1109/4235.996017}
}

@article{delamerSafePathPlanning2021,
  title = {Safe Path Planning for {UAV} Urban Operation under {GNSS} Signal Occlusion Risk},
  author = {Delamer, Jean-Alexis and Watanabe, Yoko and Chanel, Caroline P.C.},
  year = {2021},
  month = aug,
  journal = {Robotics and Autonomous Systems},
  volume = {142},
  pages = {103800},
  issn = {09218890},
  doi = {10.1016/j.robot.2021.103800}
}

@inproceedings{ghambariEnhancedNSGAIIMultiobjective2020,
  title = {An Enhanced {NSGA-II} for Multiobjective {UAV} Path Planning in Urban Environments},
  booktitle = {Proceeding of 2020 IEEE 32nd International Conference on Tools with Artificial Intelligence (ICTAI)},
  author = {Ghambari, Soheila and Golabi, Mahmoud and Lepagnot, Julien and Brevilliers, Mathieu and Jourdan, Laetitia and Idoumghar, Lhassane},
  year = {2020},
  month = nov,
  pages = {106--111},
  publisher = {IEEE},
  address = {Baltimore, MD, USA},
  doi = {10.1109/ICTAI50040.2020.00027},
  isbn = {978-1-72819-228-4}
}

@article{gongMultiobjectiveImmuneAlgorithm2008,
  title = {Multiobjective Immune Algorithm with Nondominated Neighbor-Based Selection},
  author = {Gong, Maoguo and Jiao, Licheng and Du, Haifeng and Bo, Liefeng},
  year = {2008},
  month = jun,
  journal = {Evolutionary Computation},
  volume = {16},
  number = {2},
  pages = {225--255},
  issn = {1063-6560, 1530-9304},
  doi = {10.1162/evco.2008.16.2.225}
}

@article{huRiskAssessmentModel2020,
  title = {Risk Assessment Model for {UAV} Cost-Effective Path Planning in Urban Environments},
  author = {Hu, Xinting and Pang, Bizhao and Dai, Fuqing and Low, Kin Huat},
  year = {2020},
  journal = {IEEE Access},
  volume = {8},
  pages = {150162--150173},
  issn = {2169-3536},
  doi = {10.1109/ACCESS.2020.3016118}
}

@article{linAdaptiveImmuneinspiredMultiobjective2018,
  title = {An Adaptive Immune-Inspired Multi-Objective Algorithm with Multiple Differential Evolution Strategies},
  author = {Lin, Qiuzhen and Ma, Yueping and Chen, Jianyong and Zhu, Qingling and Coello, Carlos A. Coello and Wong, Ka-Chun and Chen, Fei},
  year = {2018},
  month = mar,
  journal = {Information Sciences},
  volume = {430--431},
  pages = {46--64},
  issn = {00200255},
  doi = {10.1016/j.ins.2017.11.030}
}

@article{linHybridEvolutionaryImmune2015,
  title = {A Hybrid Evolutionary Immune Algorithm for Multiobjective Optimization Problems},
  author = {Lin, Qiuzhen and Chen, Jianyong and Zhan, Zhi-Hui and Chen, Wei-Neng and Coello Coello, Carlos and Yin, Yilong and Lin, Chih-Min and Zhang, Jun},
  year = {2015},
  journal = {IEEE Transactions on Evolutionary Computation},
  pages = {711--729},
  issn = {1089-778X, 1089-778X, 1941-0026},
  doi = {10.1109/TEVC.2015.2512930}
}

@inproceedings{liuEvolutionaryApproachMultiparty2020,
  title = {Evolutionary Approach to Multiparty Multiobjective Optimization Problems with Common {P}areto Optimal Solutions},
  booktitle = {Proceeding of 2020 IEEE Congress on Evolutionary Computation (CEC)},
  author = {Liu, Wenjie and Luo, Wenjian and Lin, Xin and Li, Miqing and Yang, Shengxiang},
  year = {2020},
  month = jul,
  pages = {1--9},
  publisher = {IEEE},
  address = {Glasgow, United Kingdom},
  doi = {10.1109/CEC48606.2020.9185747},
  isbn = {978-1-72816-929-3}
}

@inproceedings{narkus-kramerFutureDemandBenefits2017,
  title = {Future Demand and Benefits for Small Unmanned Aerial Systems ({UAS}) Package Delivery},
  booktitle = {Proceeding of 17th AIAA Aviation Technology, Integration, and Operations Conference},
  author = {{Narkus-Kramer}, Marc P.},
  year = {2017},
  month = jun,
  publisher = {American Institute of Aeronautics and Astronautics},
  address = {Denver, Colorado},
  doi = {10.2514/6.2017-4103},
  isbn = {978-1-62410-508-1}
}

@article{pangUAVPathOptimization2022a,
  title = {{UAV} Path Optimization with an Integrated Cost Assessment Model Considering Third-Party Risks in Metropolitan Environments},
  author = {Pang, Bizhao and Hu, Xinting and Dai, Wei and Low, Kin Huat},
  year = {2022},
  month = jun,
  journal = {Reliability Engineering \& System Safety},
  volume = {222},
  pages = {108399},
  issn = {09518320},
  doi = {10.1016/j.ress.2022.108399}
}

@article{pengConstrainedMultiObjectiveOptimization2022,
  title = {Constrained Multi-Objective Optimization for {UAV}-Enabled Mobile Edge Computing: Offloading Optimization and Path Planning},
  author = {Peng, Chaoda and Huang, Xumin and Wu, Yuan and Kang, Jiawen},
  year = {2022},
  month = apr,
  journal = {IEEE Wireless Communications Letters},
  volume = {11},
  number = {4},
  pages = {861--865},
  issn = {2162-2337, 2162-2345},
  doi = {10.1109/LWC.2022.3149007}
}

@article{pengDecompositionbasedConstrainedMultiobjective2022,
  title = {A Decomposition-Based Constrained Multi-Objective Evolutionary Algorithm with a Local Infeasibility Utilization Mechanism for {UAV} Path Planning},
  author = {Peng, Chaoda and Qiu, Shaojian},
  year = {2022},
  month = mar,
  journal = {Applied Soft Computing},
  volume = {118},
  pages = {108495},
  issn = {15684946},
  doi = {10.1016/j.asoc.2022.108495}
}

@incollection{sheNewEvolutionaryApproach2021,
  title = {A New Evolutionary Approach to Multiparty Multiobjective Optimization},
  booktitle = {Advances in Swarm Intelligence},
  author = {She, Zeneng and Luo, Wenjian and Chang, Yatong and Lin, Xin and Tan, Ying},
  editor = {Tan, Ying and Shi, Yuhui},
  year = {2021},
  volume = {12690},
  pages = {58--69},
  publisher = {Springer International Publishing},
  address = {Cham},
  doi = {10.1007/978-3-030-78811-7_6},
  isbn = {978-3-030-78810-0 978-3-030-78811-7}
}

@article{sunEvolutionaryAlgorithmConstraint2022,
  title = {An Evolutionary Algorithm With Constraint Relaxation Strategy for Highly Constrained Multiobjective Optimization},
  author = {Sun, Zhichao and Ren, Hang and Yen, Gary G. and Chen, Tianfu and Wu, Junjie and An, Hongyang and Yang, Jianyu},
  year = {2022},
  journal = {IEEE Transactions on Cybernetics},
  pages = {1--15},
  issn = {2168-2267, 2168-2275},
  doi = {10.1109/TCYB.2022.3151974}
}

@article{sunPathPlanningGEOUAV2016,
  title = {Path Planning for {GEO-UAV} Bistatic SAR Using Constrained Adaptive Multiobjective Differential Evolution},
  author = {Sun, Zhichao and Wu, Junjie and Yang, Jianyu and Huang, Yulin and Li, Caipin and Li, Dongtao},
  year = {2016},
  month = nov,
  journal = {IEEE Transactions on Geoscience and Remote Sensing},
  volume = {54},
  number = {11},
  pages = {6444--6457},
  issn = {0196-2892, 1558-0644},
  doi = {10.1109/TGRS.2016.2585184}
}

@article{tianPlatEMOMATLABPlatform2017,
	title={{PlatEMO}: A MATLAB platform for evolutionary multi-objective optimization [educational forum]},
	author={Tian, Ye and Cheng, Ran and Zhang, Xingyi and Jin, Yaochu},
	journal={IEEE Computational Intelligence Magazine},
	volume={12},
	number={4},
	pages={73--87},
	year={2017},
	publisher={IEEE}
}

@article{torijaEffectsHoveringUnmanned2020,
  title = {Effects of a Hovering Unmanned Aerial Vehicle on Urban Soundscapes Perception},
  author = {Torija, Antonio J. and Li, Zhengguang and Self, Rod H.},
  year = {2020},
  month = jan,
  journal = {Transportation Research Part D: Transport and Environment},
  volume = {78},
  pages = {102195},
  issn = {13619209},
  doi = {10.1016/j.trd.2019.11.024}
}

@article{zhangMultiobjectiveParticleSwarm2022,
  title = {Multi-Objective Particle Swarm Optimization with Multi-Mode Collaboration Based on Reinforcement Learning for Path Planning of Unmanned Air Vehicles},
  author = {Zhang, Xiangyin and Xia, Shuang and Li, Xiuzhi and Zhang, Tian},
  year = {2022},
  month = aug,
  journal = {Knowledge-Based Systems},
  volume = {250},
  pages = {109075},
  issn = {09507051},
  doi = {10.1016/j.knosys.2022.109075}
}

@article{kwong2020tree,
  title={Tree height mapping and crown delineation using {LiDAR}, large format aerial photographs, and unmanned aerial vehicle photogrammetry in subtropical urban forest},
  author={Kwong, Ivan HY and Fung, Tung},
  journal={International Journal of Remote Sensing},
  volume={41},
  number={14},
  pages={5228--5256},
  year={2020},
  publisher={Taylor \& Francis}
}

@article{torija2021psychoacoustic,
  title={A psychoacoustic approach to building knowledge about human response to noise of unmanned aerial vehicles},
  author={Torija, Antonio J and Clark, Charlotte},
  journal={International Journal of Environmental Research and Public Health},
  volume={18},
  number={2},
  pages={682},
  year={2021},
  publisher={MDPI}
}

@article{butilua2022urban,
  title={Urban Traffic Monitoring and Analysis Using Unmanned Aerial Vehicles ({UAVs}): A Systematic Literature Review},
  author={Butil{\u{a}}, Eugen Valentin and Boboc, R{\u{a}}zvan Gabriel},
  journal={Remote Sensing},
  volume={14},
  number={3},
  pages={620},
  year={2022},
  publisher={MDPI}
}

@article{outay2020applications,
  title={Applications of unmanned aerial vehicle ({UAV}) in road safety, traffic and highway infrastructure management: Recent advances and challenges},
  author={Outay, Fatma and Mengash, Hanan Abdullah and Adnan, Muhammad},
  journal={Transportation research part A: policy and practice},
  volume={141},
  pages={116--129},
  year={2020},
  publisher={Elsevier}
}

@article{zitzler1999multiobjective,
	title={Multiobjective evolutionary algorithms: a comparative case study and the strength Pareto approach},
	author={Zitzler, Eckart and Thiele, Lothar},
	journal={IEEE transactions on Evolutionary Computation},
	volume={3},
	number={4},
	pages={257--271},
	year={1999},
	publisher={IEEE}
}

@inproceedings{chang2022multiparty,
  title={Multiparty Multiobjective Optimization By {MOEA/D}},
  author={Chang, Yatong and Luo, Wenjian and Lin, Xin and She, Zeneng and Shi, Yuhui},
  booktitle={Proceeding of 2022 IEEE Congress on Evolutionary Computation (CEC)},
  pages={01--08},
  year={2022},
  organization={IEEE}
}

@article{zhu2022ucav,
  title={UCAV path planning for avoiding obstacles using cooperative co-evolution spider monkey optimization},
  author={Zhu, Haoran and Wang, Yunhe and Li, Xiangtao},
  journal={Knowledge-Based Systems},
  volume={246},
  pages={108713},
  year={2022},
  publisher={Elsevier}
}

@article{xu2023cooperative,
  title={Cooperative path planning optimization for multiple UAVs with communication constraints},
  author={Xu, Liang and Cao, Xianbin and Du, Wenbo and Li, Yumeng},
  journal={Knowledge-Based Systems},
  volume={260},
  pages={110164},
  year={2023},
  publisher={Elsevier}
}
\vspace{-1.05 cm}
\vfill
\end{document}